# EVALUATION OF DRAIN, A DEEP-LEARNING APPROACH TO RAIN RETRIEVAL FROM GPM PASSIVE MICROWAVE RADIOMETER.

Nicolas Viltard, Vibolroth Sambath, Pierre Lepetit*, Audrey Martini, Laurent Barthès, Cécile Mallet

LATMOS-IPSL, Université Paris-Saclay, UVSQ, CNRS, 78280, Guyancourt, France
*Météo-France, Avenue Coriolis, Toulouse

*Abstract*— **Retrieval of rain from Passive Microwave radiometers data has been a challenge ever since the launch of the first Defense Meteorological Satellite Program in the late 70s. Enormous progress has been made since the launch of the Tropical Rainfall Measuring Mission (TRMM) in 1997 but until recently the data were processed pixel-by-pixel or taking a few neighboring pixels into account. Deep learning has obtained remarkable improvement in the computer vision field, and offers a whole new way to tackle the rain retrieval problem. The Global Precipitation Measurement (GPM) Core satellite carries similarly to TRMM, a passive microwave radiometer and a radar that share part of their swath. The brightness temperatures measured in the 37 and 89 GHz channels are used like the RGB components of a regular image while rain rate from Dual Frequency radar provides the surface rain. A U-net is then trained on these data to develop a retrieval algorithm: Deep-learning RAIN (DRAIN). With only four brightness temperatures as an input and no other a priori information, DRAIN is offering similar or slightly better performances than GPROF, the GPM official algorithm, in most situations. These performances are assumed to be due to the fact that DRAIN works on an image basis instead of the classical pixel-by-pixel basis.**

## I. MOTIVATION

We have developed a rain retrieval algorithm, DRAIN, based on the use of deep-learning techniques. The architecture and principles driving DRAIN are presented in more detail in [1] but, namely, a database of co-located rain rates from the Dual frequency Precipitation Radar (DPR) and brightness temperatures from the Global Precipitation Measurement Microwave radiometer (GPM-GMI) were fed into a U-net [2]. This type of convolutional network was successfully used to de-clutter radar images [3] and thus appeared to be well adapted to detect the contours of rain events.

The main difference between more classical retrieval algorithms, either Bayesian-based [4, 5] or machine learning-based [6-12], and DRAIN arises from the fact that the latter processes the data as images while the formers work on a pixel-by-pixel basis.

In the present paper, we will not go into the details of DRAIN which have already been presented in [1], but we will focus on a more thorough validation of a more mature version of the algorithm. The two main improvements that must be highlighted between the initial version presented in [1] and the current version are described hereafter. First, the database was increased from about 52,000 images to about 103,000 allowing us to build a training database of 70,000 images for training and 33,000 images for validation. Data from the whole years 2014 to 2018 and a few months from 2020 and 2021 are used but the whole year 2019 was kept separate for the performance assessment (test) and most results presented hereafter are computed for that year. This large database is meant to dampen the effects of seasonal and interannual variability of rain.

Second, DRAIN retrieves now a set of 99 quantiles instead of a simple averaged rain rate as in [1]. These quantiles represent the probability that the rain rate is below a certain threshold. Hereafter, when unspecified, the quantile 50 % (median) is used as rain proxy. The loss function for quantile regression is the one proposed in [13]. Retrieving quantile for rain is interesting because as shown hereafter, it is possible to infer a confidence interval for the results. It would also be possible from the retrieved Cumulative Distribution Function (CDF) of rain intensity to deduce a Probability Density Function (PDF, not presented here).

Section II will give a short presentation of the database construction while section III gives some elements about the methodology and the associated cost function. Section IV offers a detailed validation of DRAIN rain rate against DPR and GPROF for the year 2019. Section V presents the comparison for the same year with Meteo-France five-minute 1 kmx1km rain mosaic. Finally, section VI presents the conclusions and perspectives.

## II. DATABASE

GMI is a conically-scanning radiometer with channels at 10.65, 18.7, 23.8, 36.6, 89.0, 166.0, 183.3+/-3 and 183+/-7 GHz. All the channels are measured in both Horizontal (H) and Vertical (V) polarization except for 23.8 and the two 183.3 GHz sounding channels (V only). In the present study, only the two 36.6 (hereafter noted 37 GHz) and 89.0 GHz channels were used. This choice was driven by the idea that most conical-scanning passive microwave radiometers for rain retrieval do have channels in the 37 and 89 GHz region thus making a transposition of DRAIN to other platforms potentially easier. These two frequencies were also selected despite being mainly affected by ice scattering, because they offer a good spatial resolution with well-defined horizontal gradients. For the GMI, the pixel resolutions are respectively: 15.6x9.4 km$^2$ for the 37 GHz and 7.2x4.4 km$^2$ for 89 GHz [14].

Since [1], some tests (not shown) were made to check if the addition of the 19 GHz channels would improve the performances of DRAIN but the results were unconclusive.

DPR surface rain product results of the merged use of the Ku- (13.4 GHz) and Ka-band (35.5 GHz) radars. DPR measures a three-dimensional reflectivity field with vertical resolution of 250 m, a horizontal resolution of 5 km, and swath widths of 245 km [14].

The DPR and GMI pixels are co-located spatially and temporally assuming that the effective one-minute lag between the two observations is negligible at the considered spatial resolution. To perform the co-location, the surface rain (precipRateESurface) of the DPR pixels falling within 5 km of a GMI pixel center position are averaged. On average, three to four DPR pixels fall into the 5 km-radius.

We will hereafter refer to training and validation database as the data used to adjust the weights and hyperparameters of the network and the test database as the one used for generalization and assessment of performances.

Because natural rain occurrence is low, a scene selection is made as follow: images with at least 100 pixels with rain > 0.1 mm.hr$^{-1}$ or at least 10 pixels > 100 mm.hr$^{-1}$ are kept to build the training/validation database. Each image is composed of 4 channels: the 37 and 89 in both H and V polarization while the surface rain is used as target.

## III. METHOD

The architecture of the U-net used for DRAIN is described in [2] (and [1]). As stated previously, this U-net was successfully used for detection and restoration of clutter echoes in weather radar raw data [3].

Weights, optimization method (Adam, [15]) and initial learning rate ($10^{-4}$) are set to the default values found in literature [16]. The trained U-net has about ~15 million parameters to be adjusted through the training phase.

The loss function used here is the classical one used for quantile regression (e.g. [13]) where the loss function is given by:

$$L_{q_j}(y_i, \hat{y}_i) = \begin{cases} q_j(y_i - \hat{y}_i) & if\ y_i - \hat{y}_i \geq 0 \\ (q_j - 1)(y_i - \hat{y}_i) & if\ y_i - \hat{y}_i < 0 \end{cases}$$

$$L_{q_j}(y, \hat{y}) = \frac{1}{N}\sum_i L_{q_j}(y_i, \hat{y}_i)$$

$$L(y, \hat{y}) = \sum_j L_{q_j}(y, \hat{y})$$

where $\hat{y}$ is the prediction, $y$ is the target, $q_j$ is the jth quantile to be estimated and N the number of pixels.

Besides the fact that retrieving quantiles gives access to more information than the mere retrieved average rain rate, the quantile regression is of particular interest here because of its demonstrated robustness to outliers. In the present study, we chose to retrieve percentiles because computation of the rain intensity probability density function is easier to compute through simple numerical derivation.

The network presented here results in a 460-epoch training on the learning and validation bases described above. As a base value for the retrieved rain rate, we will use hereafter the median which correspond to the 50$^{th}$ quantile.

Because the chosen architecture is made of a 16-level of filtering, the retrieval image sizes have to be multiple of 16. For the moment the retrieved images are thus limited to 208 pixels per 2960 scan from the original 221 pixels and close to 2963 scans (the effective number of scans may vary slightly from one granule to the next but remains larger than 2960).

## IV. COMPARISON WITH DPR

Two example cases, excluded from the learning and validation dataset, are given on Fig. 2 and 3. The first case is super typhoon Nanmadol observed on September 16$^{th}$, 2022, while it is close to its peak intensity over the Philippine Sea, South of Japan. The second case is a frontal band over France observed on August 18$^{th}$, 2018. The first case is a good illustration of oceanic retrieval while the second case is a mixed continental and coastal case.

In addition to comparisons with the DPR in the common part of the swath, comparisons with the GPROF algorithm [4] are also presented. GPROF is the GPM operational product, based on a per-pixel-Bayesian approach. GPROF is considered as a reference in the GPM community and uses auxiliary data to constrain the solutions (temperature, surface type, humidity, cloud cover…).

Note that for all the comparisons presented hereafter with DPR and GPROF and in part V also, we use the 50$^{th}$ quantile (median) as our best estimator for DRAIN. For both situations, DRAIN and DPR are qualitatively very close and resemble GPROF except that the latter exhibits a lot of light rain pixels (< 0.2-0.3 mm/hr). The DPR algorithm is estimated to have a minimal detection threshold near 0.2 mm.hr$^{-1}$ [17]. Because we perform a spatial average over two to four DPR pixels, DRAIN is expected to have a theoretical detection threshold between 0.03 and 0.1 mm.hr$^{-1}$ depending on the number of averaged pixels.

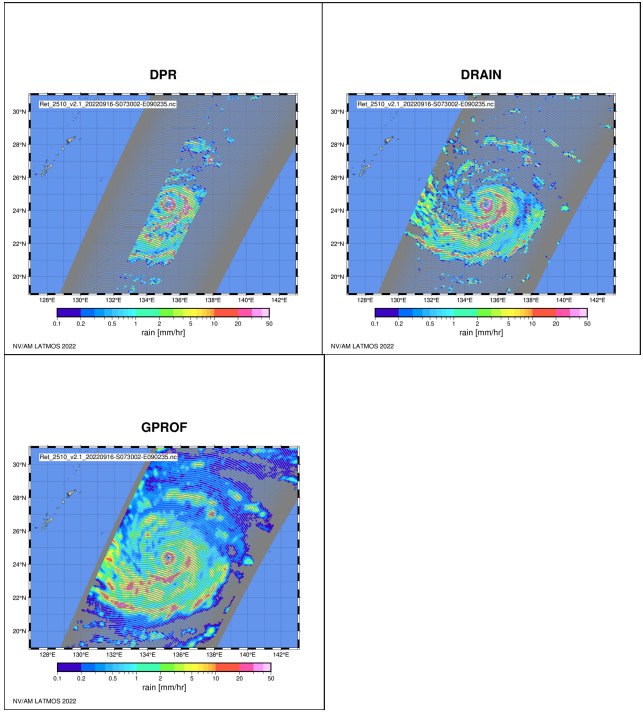

Figure 2: surface rain for super Typhoon Nanmadol (16W) on 16th September 2022 at 08:00 UTC. DPR at DRAIN resolution (top left), DRAIN (top right) and GPROF (Bottom left).

The most noticeable difference between GPROF and DRAIN is particularly visible on Fig. 2 (super typhoon Nanmadol) where the GPROF retrieves many more rainy pixels with intensities below 0.4 mm/hr than both DRAIN and DPR. DRAIN on this aspect is very consistent with the DPR as one would expect since DRAIN was developed solely from DPR data without any other assumptions concerning the relation brightness temperatures – rain intensities. The reasons behind this light rain enhancement in GPROF is described in [17]. This light rain difference is more likely to be visible over ocean where, in addition to the higher frequency channels, GPROF uses 10 and 19 GHz channels which are directly linked to liquid water emission.

The second comparison is made at global level over the whole year 2019. A few granules with bad $T_{BS}$ at either 37 or 89 GHz are excluded. A mask is used to check systematic differences between land and ocean. Unlike for GPROF, this mask is not used as an input in DRAIN but only once the retrieval is performed to assess the performances over the two different surface types. This mask is the python "regionmask" package which is based on Natural Earth, a free vector and raster map data (naturalearthdata.com) at 1:110m resolution (1 cm = 1100 km).

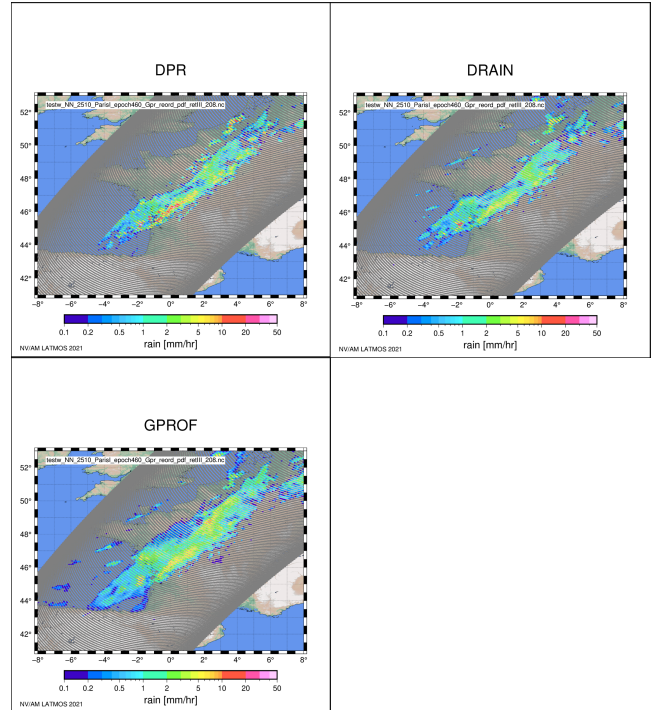

Figure 3: same as Fig. 2 but for a frontal rain band over France on 18th of August 2018.

Table 1 and 2 give respectively the contingency table for OCEAN (approximately 530 million pixels) and LAND (approximately 218 million pixels). DRAIN is well-balanced between False Alarm (FA) and Bad Detection (BD) occurrence although the False Alarm has a high RMSE of 10.77 mm/hr (17.36 above ocean and 7.70 above land). The high probability of FA for GPROF are mostly due to the light rain treatment of the latter, as mentioned above. On the other hand, these FA have a very small RMSE of 0.30 mm/hr because they are made of very light rain rates.

| OCEAN DPR | DRAIN Rain | No Rain | GPROF Rain | No Rain |
|---|---|---|---|---|
| Rain | 6.01 | 1.97 | 5.93 | 2.05 |
| No Rain | 1.23 | 90.79 | 13.14 | 78.88 |

Table 1: Contingency in % table for OCEAN pixels on the whole dataset of 2019.

| LAND DPR | DRAIN Rain | No Rain | GPROF Rain | No Rain |
|---|---|---|---|---|
| Rain | 3.49 | 1.26 | 3.46 | 1.29 |
| No Rain | 0.91 | 94.34 | 10.44 | 84.81 |

Table 2: same as Table 1 but for LAND pixels.

|        | OCEAN |      | LAND |      |
|--------|-------|------|------|------|
|        | POD   | FAR  | POD  | FAR  |
| DRAIN  | 0.75  | 0.17 | 0.74 | 0.21 |
| GPROF  | 0.74  | 0.69 | 0.72 | 0.75 |

Table 3: Probability of Detection (POD) and False Alarm Ratio (FAR) over land and ocean for DRAIN and GPROF, when comparing with DPR in the common part of the swath.

|        | DRAIN |      | GPROF |      |
|--------|-------|------|-------|------|
|        | Bias  | RMSE | Bias  | RMSE |
| LAND   | 0.31  | 2.67 | -0.24 | 3.39 |
| OCEAN  | 0.26  | 2.98 | 0.08  | 3.18 |
| TOTAL  | 0.27  | 2.92 | 0.01  | 3.22 |

Table 4: The numbers are computed on the respective true positive with a threshold of $10^{-4}$ mm/hr. Bias, in mm/hr are respectively DPR-DRAIN and DPR-GPROF.

The BD for both DRAIN and GPROF show an average at 0.33 mm/hr and 0.50 mm/hr respectively, meaning it is mainly made of the detection of some light rain by the DPR.

Table 3 shows the Probability Of Detection (POD) and the False Alarm Ratio (FAR) which are often used to assess the performances of rain retrieval algorithms [18]. Perfect POD is 1 while perfect FAR is 0. Both algorithms offer similar PODs with a small advantage for DRAIN, specially over land. The low score of GPROF's FAR is logically due to the light rain over-detection with respect to DPR as mentioned above.

Table 4 shows the bias and Root Mean Square Error (RMSE) for the two algorithms, compared to DPR for their respective true positive only. A $10^{-4}$ mm/hr threshold is applied to make sure that no random numerical noise will contaminate the results. GPROF is always better in terms of bias but DRAIN is better in terms of RMSE. This is supported by Fig 4 that shows the corresponding scatter plots for rain intensities between 0 and 100 mm/hr. DRAIN has been trained with the DPR so, once again it is expected that its performances with respect to the latter will be optimal.

A substantial spread is however observed which is expected due to the potential parallax effects between the GMI TBs and the DPR surface rain which cannot be totally compensated. Over both land and ocean, systematic underestimation starts appearing at about 20 mm/hr and increases as the DPR rain rate increases.

Over ocean, GPROF shows similar features, slightly exacerbated for DPR rain rates above 25 mm/hr. Over land, comparison is made difficult by the fact that the GPROF rain rate is made of the Ku estimates completed with NOAA's Multi-Radar/Multi-Sensor system (MRMS) [19].

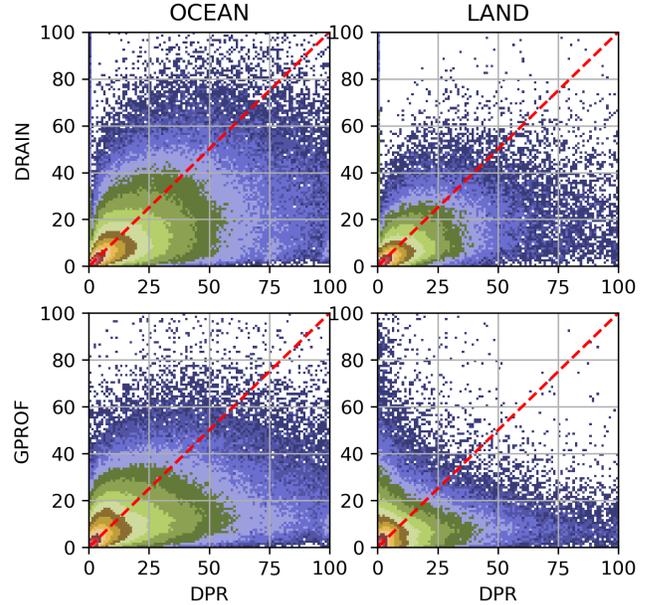

Figure 4: scatter plot of DRAIN (top row) and GPROF (bottom row) against DPR for OCEAN pixels (left hand side column) LAND pixels (right hand side column) for the whole of 2019. Colors show the density of point from red (densest) to blue (least dense). The red dashed line on each of the graph is the x=y line.

Fig. 5 shows the histograms of DRAIN, GPROF and DPR for the light rain rates, emphasizing the higher probability of rain below 0.25 mm/hr proposed by GPROF when compared to the two other estimators. This is true for land and even more for ocean. It is to be once again mentioned that these light rain rates have been enhanced in GPROF's a priori database, based on Cloudsat statistics [19] to compensate in particular for the Ku-DPR detection threshold (~12 dBZ).

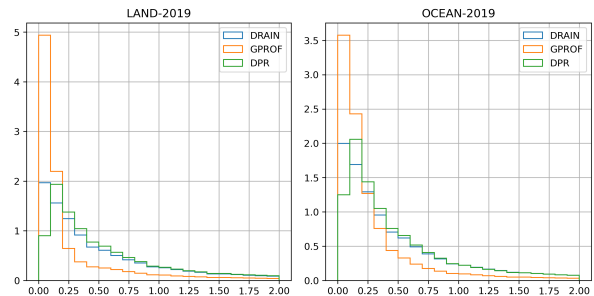

Figure 5: pdf of rain intensity for DPR, GPROF and DRAIN for the year 2019, focusing on the light rain between 0 and 2 mm.hr$^{-1}$.

| Rain interval mm/hr | 50% | 90% |
|---|---|---|
| 0 to 0.1 | 73.96 | 97.03 |
| 0.1 to 1. | 56.36 | 93.93 |
| 1 to 10 | 46.49 | 87.01 |
| 10 and above | 33.65 | 75.47 |
| All | 53.96 | 91.78 |

Table 5: retrieved confidence intervals as a function of rain intensities.

Since percentiles are retrieved by the network, among other applications, it is possible to define confidence intervals for the retrieved rain intensity. The rain rate, $RR_j$, given for the j$^{th}$ percentile means that the a-priori probability that the DPR rain rate is between 0 and $RR_j$ is j%. In Table 5 the 50% and 90% confidence interval are verified against the DPR surface rain. For the true positive, pixels for which the DPR rain rates falls indeed between percentiles 25 and 75% and 5 and 95% are counted. It can be seen that overall, the confidence interval is indeed reliable but this actually depends on the rain interval under consideration. Up to about 10 mm/hr, the confidence interval is robust but above, the systematic underestimation of the retrieved rain rate degrades the results.

Figure 6 is a zoom for rain rates between 0 and 50 mm/hr of DRAIN vs DPR. The scatter plot is not differentiating land and ocean. On top of the scatter plot is the 90 % confidence (shade of blue) and the 50 % confidence (shade of brown). These confidences are computed using the 5$^{th}$ and 95$^{th}$ retrieved percentiles and the 25$^{th}$ and 75$^{th}$ retrieved percentiles respectively.

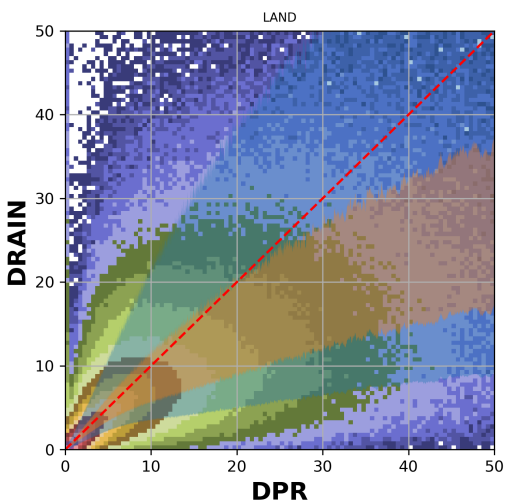

*Figure 6: 2-D histogram of the retrieved rain vs. initial (DPR) rain, color indicates density of points. Blue-shaded area is the 90 % confidence interval and brow-shaded area is the 50 % confidence interval.*

Figure 7 shows these differences between GPROF, DRAIN and DPR on a 1°x1° global map. The three estimators are first averaged on 1-degree squares, keeping only the pixels above 10$^{-3}$ mm/hr.

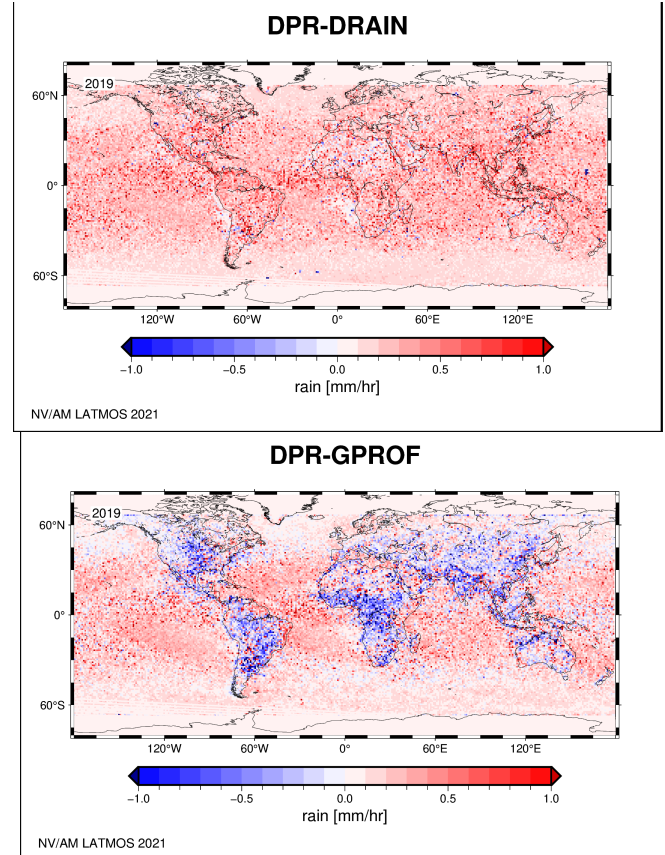

*Figure7: 1°x1° averaged difference between top: DPR -DRAIN and bottom DPR - GPROF. The differences are performed at pixel level and then averaged and this, over all the orbits of 2019.*

DRAIN offers a very consistent bias overall. No difference is noticeable between land and ocean or artifact over coastal area which is always very difficult to handle in such retrieval problems. However, the bias appears to be mostly positive (underestimation) while a more balanced distribution of positive and negative values was expected. This might be due to the choice of the median quantile as the rain estimator, which is somewhat arbitrary.

A slight latitude dependence of the error can be observed as the error seems smaller above 50 N and below 50 S over ocean and slightly larger in the ITCZ. Errors appear to be also larger above mountainous areas like the Tibetan plateau, the Rocky Mountains and somewhat the Andes.

GPROF's error range is very similar to DRAIN's but with a marked difference between land and ocean which was already mentioned. A slight dependence to the latitude can also be observed but less marked than DRAIN.

## V. RESULTS ON MÉTÉO-FRANCE MOSAIC

An assessment of the performances was also conducted using Météo-France five-minute mosaic product which is a good reference for mid-latitude QPE. The used product is described in [20] and comes as a 1536x1536 pixels grid of 1 km resolution every 5 minutes over the whole of year 2019. Co-location in time was performed by matching the closest mosaic in time with the "middle" time of GMI overpass (the GMI overpass lasts about 3 minutes). The accumulation over 5 minutes is converted in mm.hr$^{-1}$ by simple multiplication by a factor 12. A quality flag is associated with the Météo-France product ranging from 0 (unreliable) to 100 % (very reliable). After visual comparison on a series of cases, it appeared that a threshold of 80 % reliability should be applied in order to eliminate spurious rain estimates particularly in the mountainous areas.

The mosaic data are then co-located and averaged at the DRAIN resolution and pixels position. Between January 1st 2019 and December 31st 2019, 1565 overpasses are kept, with the condition that more than 50 DRAIN pixels fall into the mosaic domain: 8° West to 12° East and 39° North to 54° North.

First, a pixel-by-pixel performance is evaluated for the three rain estimators: DRAIN, DPR and GPROF. Contingency table and F1-scores are computed and presented Table 6, 7 and 8 respectively. The total number of pixels taken into account differs for each estimator because of the swath difference. The three estimators show performances that are close with a few differences. GPROF shows a better POD than both DRAIN and DPR but its FAR is degraded by the excess of light rain detected. On the other hand, DPR and DRAIN miss some of the rain which degrades their respective POD but their precision remains high.

| Ref.\DRAIN | Positive | Negative | POD |
|---|---|---|---|
| Positive | 4.85 % | 9.81 % | 0.33 |
| Negative | 0.36 % | 84.98 % | FAR |
| Precision | 0.93 | | 0.07 |
| F1-score | 0.49 | | |

Table 6: Contingency table and F1-score for DRAIN with Météo-France mosaic as a reference. The total number of co-located pixels is 6645997.

| Ref.\DPR | Positive | Negative | POD |
|---|---|---|---|
| Positive | 5.59 % | 8.53 % | 0.40 |
| Negative | 0.44 % | 84.44 % | FAR |
| Precision | 0.93 | | 0.07 |
| F1-score | 0.56 | | |

Table 7: same as Table 4 but for DPR. The total number of co-located pixels is 1434964.

| Ref.\GPROF | Positive | Negative | POD |
|---|---|---|---|
| Positive | 7.38 % | 7.26 % | 0.50 |
| Negative | 4.49 % | 80.88 % | FAR |
| Precision | 0.62 | | 0.38 |
| F1-score | 0.56 | | |

Table 8: same as Table 4 but for GPROF. The total number of co-located pixels is 5611805.

The DRAIN, DPR and GPROF data are also averaged on 0.2°x0.2° grid to minimize the possible impact of localization errors between the ground and satellite-based estimates. Figure 8 shows the scatter plot for each of the estimators against the Météo-France mosaic. Only the grid-boxes where the mosaic >= 0 mm.hr$^{-1}$ are accounted for. For each scatter plot, the associated linear regression (blue-dotted line) is shown. For DRAIN the corresponding R$^2$=0.29 while it is 0.10 for DPR and 0.20 for GPROF. Below 1 mm/hr, all estimators are mostly centered on the x=y line and then the points spread out with a general underestimation. DPR appears more spread than both DRAIN and GPROF but the data sample is about four times smaller and the narrow swath might induce more important edges artifacts. GPROF appears to show a slight underestimation (~0.1 mm/hr) around 0.6-0.7 mm/hr, in the densest part of the scatter plot. Globally however, GPROF and DRAIN appear to offer similar performances.

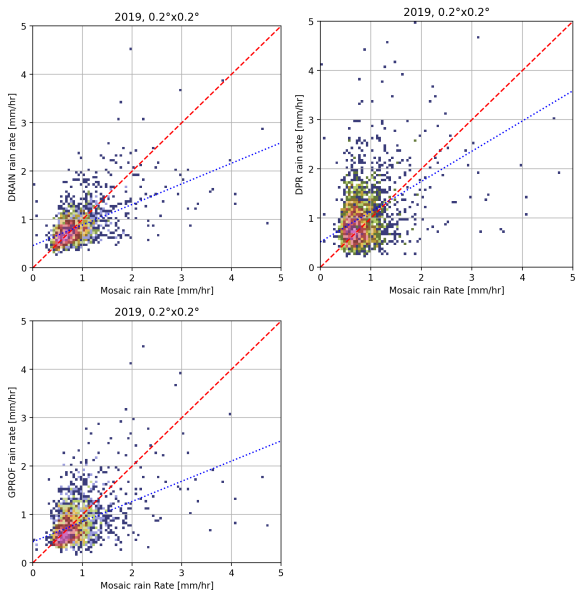

*Figure 8: Scatter plot of DRAIN (top left), DPR (top left) and GPROF (bottom left) vs Météo-France mosaic over all the overpasses of 2019. In each graph, red-dashed line is the x=y line, blue-dotted line is the linear regression and the color are proportional to the density of points (from purple: densest to blue: least dense).*

Figure 9 shows the spatial distribution of differences Mosaic-DRAIN and Mosaic-GPROF, in mm.hr$^{-1}$, for 2019. Except for a few grid-boxes, the difference remains between -2 and +2 mm.hr$^{-1}$. The general pattern and amplitude of the difference are similar between the two estimators.

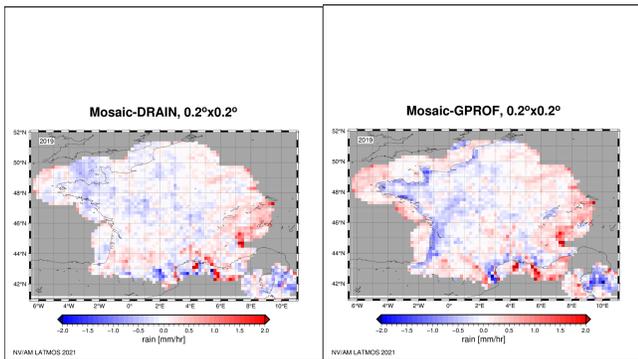

*Figure 9: Difference between Météo-France mosaic and DRAIN on the left hand side and GPROF on the right hand side. The difference is computed over 0.2°x0.2° boxes over the whole year 2019.*

In the mountainous area of the Alps and somewhat on the Mediterranean shores, the errors are almost identical on a surprising number of boxes which might show a problem with the Météo-France estimate. It is noticeable though, that GPROF has a very clear artifact of overestimating the land part of the coastal regions. This is likely due to the difference between the land and the ocean version of GPROF. Most of DRAIN error structure appears to be much more random and better balanced, yet there is continuity from one box to the next, showing that the errors are not pure noise. This spatial continuity of the errors is also true for GPROF.

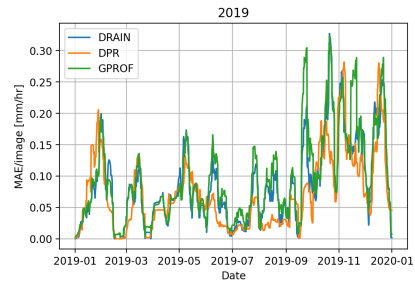

*Figure 10: Mean Average Error between the DRAIN, GPROF and DPR as a function of time over 2019.*

Figure 10 shows the dependance with time of the Mean Average Error for the three estimators. Environmental conditions change depending on the season and this is eventually even more pronounced over land with possible snow cover on the ground. All three estimators follow more or less the same patterns with an increase in the errors during winter. The DPR should be considered with some care because its coverage is substantially lower than both DRAIN and GPROF inducing some possible representativeness artifacts.

## VI. CONCLUSIONS

From a set of 103,000 images of co-located data between GMI brightness temperatures and DPR surface rain, a U-net was trained to retrieve the latter over the whole swath of the radiometer. To minimize the impact of surface emissivity, work with the highest spatial resolution possible and at the same time remaining easily transposable to other instruments, only 37 and 89 GHz horizontal and vertical polarization brightness temperatures are used as an input. The strength of U-nets is their ability to process brightness temperature scenes as an image, as opposed to most existing algorithms that proceed pixel by pixel.

Evaluation of the developed algorithm is performed two ways. First, a comparison with the DPR surface rain itself on a set of images that was not in the training database (whole of 2019) is presented. DRAIN shows a good agreement in terms of structure and intensities which ensures the good quality of the generalization. When compared on the same dataset, DRAIN is generally on par with GPROF or slightly better. This is most noticeable over land where GPROF was not trained on DPR surface rain.

Second, a comparison is performed with Météo-France 1 km²-resolution mosaic over the same period as a fully independent dataset. The trends are similar. DRAIN performances are close to GPROF if not slightly better. The most noticeable difference is observed in the coastal regions where GPROF tends to overestimate the rain intensities when compared to the mosaic.

These four channels used here are present on most passive microwave radiometers of the GPM constellation which will facilitate transposition of the developed U-net to other available sensors of the constellation as proposed in [21].

## Acknowledgments

This research is work is supported by TOSCA of Centre National d'Etudes Spatiales (CNES) and INSU PNTS program. Our thanks to the reviewers for their helpful comments.